\documentclass{article} %
\usepackage{iclr2023_conference,times}

\usepackage{amsmath,amsfonts,bm}

\def\eqref#1{equation~\ref{#1}}

\def\1{\bm{1}}

\DeclareMathAlphabet{\mathsfit}{\encodingdefault}{\sfdefault}{m}{sl}
\SetMathAlphabet{\mathsfit}{bold}{\encodingdefault}{\sfdefault}{bx}{n}

\newcommand{\R}{\mathbb{R}}

\usepackage{graphicx}
\usepackage{amsmath}
\usepackage{amssymb}
\usepackage{booktabs}
\usepackage{soul}
\usepackage{xcolor}
\usepackage{multicol}
\usepackage{multirow}
\usepackage{subfig}

\usepackage{hyperref}
\usepackage{url}
\usepackage[normalem]{ulem}

\usepackage[percent]{overpic}

\newcommand{\ourmethod}{MDM}
\newcommand{\ourmethodlong}{Motion Diffusion Model}

\title{Human \ourmethodlong{}}

\author{Guy Tevet, Sigal Raab, Brian Gordon, Yonatan Shafir, \\
\textbf{Daniel Cohen-Or and Amit H. Bermano}\\
Tel Aviv University, Israel\\
\texttt{{guytevet@mail.tau.ac.il}} \\
}

\iclrfinalcopy %
\begin{document}

\maketitle

\begin{abstract}
Natural and expressive human motion generation is the holy grail of computer animation.
It is a challenging task, due to the diversity of possible motion, human perceptual sensitivity to it, and the difficulty of accurately describing it. Therefore, current generative solutions are either low-quality or limited in expressiveness. 
Diffusion models, which have already shown remarkable generative capabilities in other domains, are promising candidates for human motion due to their many-to-many nature, but they tend to be resource hungry and hard to control. 
In this paper, we introduce \ourmethodlong{} (\ourmethod{}), a carefully adapted classifier-free diffusion-based generative model for the human motion domain. 
\ourmethod{} is transformer-based, combining insights from motion generation literature. 
A notable design-choice is the
prediction of the sample, rather than the noise, in each diffusion step. This facilitates the use of established geometric losses on the locations and velocities of the motion, such as the foot contact loss. As we demonstrate, \ourmethod{} is a generic approach, enabling different modes of conditioning, and different generation tasks. We show that our model is trained with lightweight resources and yet achieves state-of-the-art results on leading benchmarks for text-to-motion and action-to-motion\footnote{Our code can be found at \url{https://github.com/GuyTevet/motion-diffusion-model}.}.
\url{https://guytevet.github.io/mdm-page/}.

\end{abstract}

\vspace{-5pt}
\section{Introduction}
\label{sec:intro}
\vspace{-5pt}

Human motion generation is a fundamental task in computer animation, with applications spanning from gaming to robotics. It is a challenging field, due to several reasons, including the vast span of possible motions, and the difficulty and cost of acquiring high quality data. For the recently emerging text-to-motion setting, where motion is generated from natural language, another inherent problem is data labeling. For example, the label "kick" could refer to a soccer kick, as well as a Karate one. At the same time, given a specific kick there are many ways to describe it, from how it is performed to the emotions it conveys, constituting  a many-to-many problem. Current approaches have shown success in the field, demonstrating plausible mapping from text to motion~\citep{petrovich22temos,tevet2022motionclip,ahuja2019language2pose}. All these approaches, however, still limit the learned distribution since they mainly employ auto-encoders or VAEs~\citep{kingma2013auto} (implying a one-to-one mapping or a normal latent distribution respectively). In this aspect, diffusion models are a better candidate for human motion generation, as they are free from assumptions on the target distribution, and are known for expressing well the many-to-many distribution matching problem we have described.

Diffusion models~\citep{sohl2015deep,song2020improved,ho2020denoising} are a generative approach that is gaining significant attention in the computer vision and graphics community. When trained for conditioned generation, recent diffusion models~\citep{ramesh2022hierarchical,saharia2022photorealistic} have shown breakthroughs in terms of image quality and semantics. The competence of these models have also been shown for other domains, including videos ~\citep{ho2022video}, and 3D point clouds ~\citep{luo2021diffusion}. 
The problem with such models, however, is that they are notoriously resource demanding and challenging to control.

\begin{figure}[t!]
\centering
\begin{overpic}[width=1\columnwidth,tics=10, trim=0mm 0 0mm 0,clip]{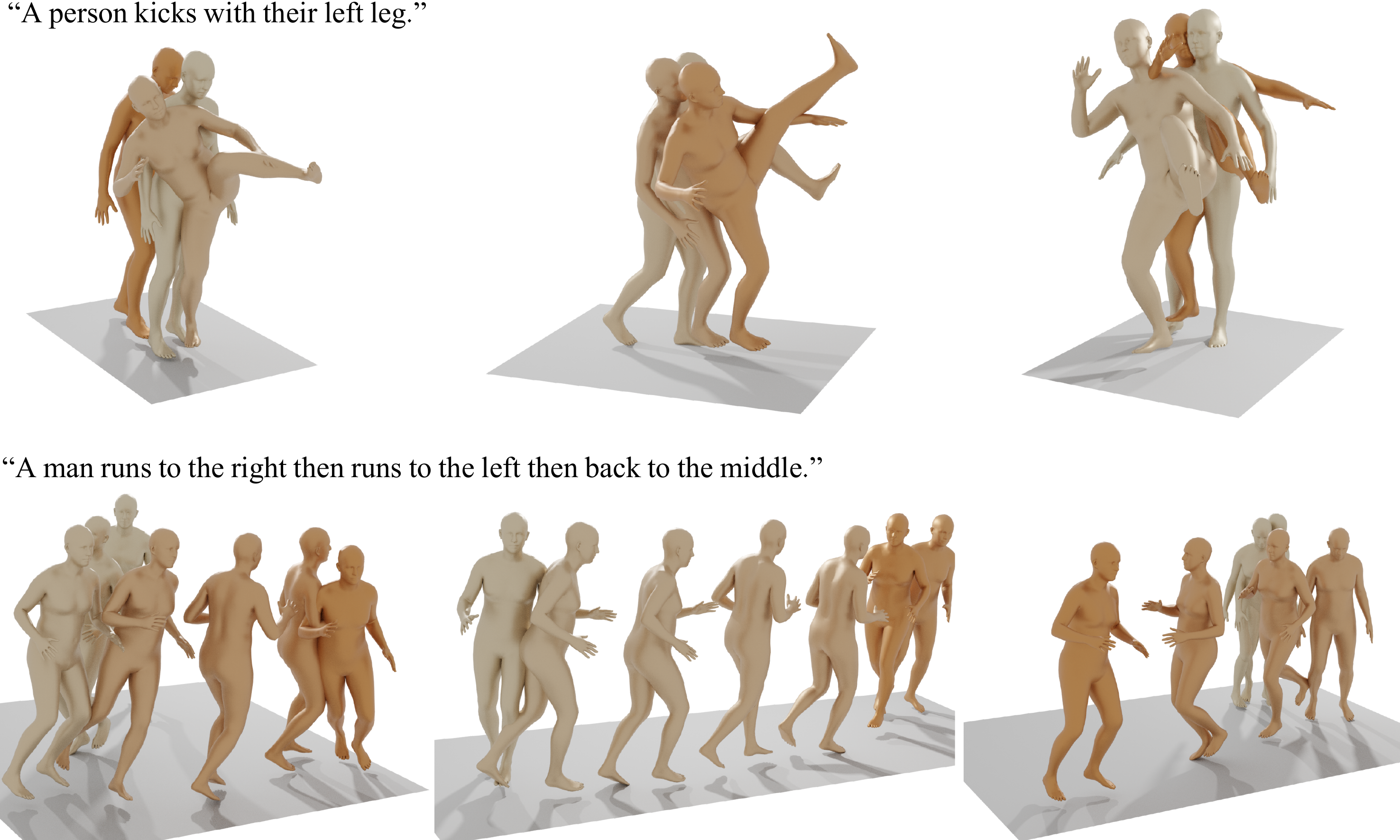}
\end{overpic}
\vspace{-5pt}
\caption{
Our \ourmethodlong{} (\ourmethod{}) reflects the many-to-many nature of text-to-motion mapping by generating diverse motions given a text prompt. Our custom architecture and geometric losses help yielding high-quality motion. Darker color indicates later frames in the sequence.
}
\vspace{-10pt}
\label{fig:teaser}
\end{figure}

In this paper, we introduce \ourmethodlong{} (\ourmethod{}) --- a carefully adapted diffusion based generative model for the human motion domain. Being diffusion-based, \ourmethod{} gains from the native aforementioned many-to-many expression of the domain, as evidenced by the resulting motion quality and diversity (Figure~\ref{fig:teaser}).
In addition, \ourmethod{} combines insights already well established in the motion generation domain, helping it be significantly more lightweight and controllable.

First, instead of the ubiquitous U-net~\citep{ronneberger2015u} backbone, \ourmethod{} is transformer-based. As we demonstrate, our architecture (Figure~\ref{fig:overview}) is lightweight and better fits the temporal and non-spatial nature of motion data (represented as a collection of joints). 
A large volume of motion generation research is devoted to learning using geometric losses~\citep{kocabas2020vibe,harvey2020robust,aberman2020skeleton}. Some, for example, regulate the velocity of the motion~\citep{petrovich21actor} to prevent jitter, or specifically consider foot sliding using dedicated terms~\citep{shi2020motionet}. 
Consistently with these works, we show that applying geometric losses in the diffusion setting improves generation.

The \ourmethod{} framework has a generic design enabling different forms of conditioning. We showcase three tasks: text-to-motion, action-to-motion, and unconditioned generation. We train the model in a classifier-free manner~\citep{ho2022classifier}, which enables trading-off diversity to fidelity, and sampling both conditionally and unconditionally from the same model.
In the text-to-motion task, our model generates coherent motions (Figure~\ref{fig:teaser}) that achieve state-of-the-art results on the HumanML3D~\citep{guo2022generating} and KIT~\citep{plappert2016kit} benchmarks. 
Moreover, our user study shows that human evaluators 
prefer our generated motions over real motions $42\%$ of the time (Figure \ref{fig:user_study_and_sweep}(a)).
In action-to-motion, \ourmethod{} outperforms the state-of-the-art \citep{guo2020action2motion,petrovich21actor}, even though they were specifically designed for this task, on the common HumanAct12~\citep{guo2020action2motion} and UESTC~\citep{ji2018large} benchmarks.

Lastly, we also demonstrate completion and editing. By adapting diffusion image-inpainting~\citep{song2020score,saharia2022palette}, we set a motion prefix and suffix, and use our model to fill in the gap. 
Doing so under a textual condition guides \ourmethod{} to fill the gap with a specific motion that still maintains the semantics of the original input. By performing inpainting in the joints space rather than temporally, we also demonstrate the semantic editing of specific body parts, without changing the others (Figure~\ref{fig:edit}). 

Overall, we introduce \ourmethodlong{}, a motion framework that achieves state-of-the-art quality in several motion generation tasks, while requiring only about three days of training on a single mid-range GPU.
It supports geometric losses, which are non trivial to the diffusion setting, but are crucial to the motion domain, and offers the combination of state-of-the-art generative power with well thought-out domain knowledge.

\vspace{-5pt}
\section{Related Work}
\label{sec:intro}
\vspace{-5pt}

\subsection{Human Motion Generation}
\vspace{-5pt}

Neural motion generation, learned from motion capture data, can be conditioned by any signal that describes the motion. Many works use parts of the motion itself for guidance. Some predict motion from its prefix poses~\citep{fragkiadaki2015recurrent,martinez2017human,hernandez2019human,guo2022back}. 
Others~\citep{harvey2018recurrent,kaufmann2020convolutional,harvey2020robust,duan2021single} solve in-betweening and super-resolution tasks using bi-directional GRU~\citep{cho2014learning} and Transformer~\citep{vaswani2017attention} architectures. 
\citet{holden2016deep} use auto-encoder to learn motion latent representation, then utilize it to edit and control motion with spatial constraints such as root trajectory and bone lengths. 
Motion can be controlled with a high-level guidance given from action class~\citep{guo2020action2motion,petrovich21actor,cervantes2022implicit}, audio~\citep{li2021dance,aristidou2022rhythm} and natural language~\citep{ahuja2019language2pose,petrovich22temos}. In most cases authors suggests a dedicated approach to map each conditioning domain into motion. 

In recent years, the leading approach for the \textit{Text-to-Motion} task is to learn a shared latent space for language and motion. JL2P~\citep{ahuja2019language2pose} learns the KIT motion-language dataset~\citep{plappert2016kit} with an auto-encoder, limiting one-to-one mapping from text to motion. TEMOS~\citep{petrovich22temos} and T2M~\citep{guo2022generating} suggest using a VAE~\citep{kingma2013auto} to map a text prompt into a normal distribution in latent space. 
Recently, MotionCLIP~\citep{tevet2022motionclip} leverages the shared text-image latent space learned by CLIP~\citep{radford2021learning} to expand text-to-motion out of the data limitations and enabled latent space editing.

The human motion manifold can also be learned without labels, as shown by \citet{holden2016deep}, V-Poser~\citep{SMPL-X:2019}, and more recently the dedicated MoDi architecture~\citep{raab2022modi}. We show that our model is capable for such an unsupervised setting as well.

\vspace{-5pt}
\subsection{Diffusion Generative Models}
\vspace{-5pt}

Diffusion models~\citep{sohl2015deep, song2020improved} are a class of neural generative models, based on the stochastic diffusion process as it is modeled in Thermodynamics. In this setting, a sample from the data distribution is gradually noised by the diffusion process. Then, a neural model learns the \textit{reverse process} of gradually denoising the sample. Sampling the learned data distribution is done by denoising a pure initial noise. \citet{ho2020denoising} and \citet{song2020denoising} further developed the practices for image generation applications. For conditioned generation, \citet{dhariwal2021diffusion}, introduced classifier-guided diffusion, which was later on adapted by GLIDE~\citep{nichol2021glide} to enable conditioning over CLIP textual representations. 
The Classifier-Free Guidance approach~\citet{ho2022classifier} enables conditioning while trading-off fidelity and diversity, and achieves better results~\citep{nichol2021glide}. In this paper, we implement text-to-motion by conditioning on CLIP in a classifier-free manner, similarly to text-to-image~\citep{ramesh2022hierarchical,saharia2022photorealistic}. Local editing of images is typically defined as an inpainting problem, where a part of the image is constant, and the inpainted part is denoised by the model, possibly under some condition~\citep{song2020score,saharia2022palette}. We adapt this technique to edit motion's specific body parts or temporal intervals (in-betweening)  according to an optional condition.

More recently, concurrent to this work, \citet{zhang2022motiondiffuse} and \citet{kim2022flame} have suggested diffusion models for motion generation. Our work requires significantly fewer GPU resources and
makes design choices that enable geometric losses, which improve results.

\vspace{-5pt}
\section{\ourmethodlong{}}
\label{sec:method}
\vspace{-5pt}

\begin{figure}[t!]
\centering
\begin{overpic}[width=\textwidth]{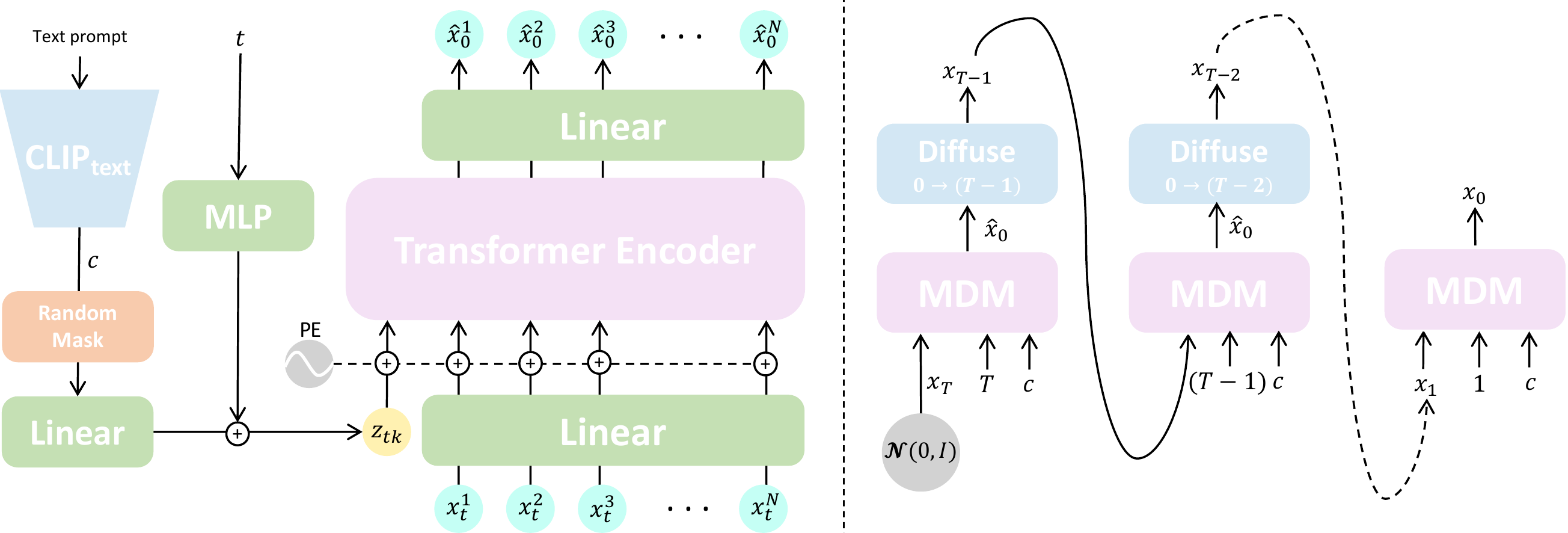}
\end{overpic}
\caption{
\textbf{(Left) \ourmethodlong{} (\ourmethod{}) overview.} The model is fed a motion sequence $x^{1:N}_{t}$ of length $N$ in a noising step $t$, as well as $t$ itself 
and a conditioning code $c$. 
$c$, a CLIP~\citep{radford2021learning} based textual embedding in this case, is first randomly masked for classifier-free learning and then projected together with $t$ into the input token $z_{tk}$. In each sampling step, the transformer-encoder predicts the final clean motion $\hat{x}^{1:N}_{0}$.
\textbf{(Right) Sampling \ourmethod{}.} Given a condition $c$, we sample random noise $x_T$ at the dimensions of the desired motion, then iterate from $T$ to $1$. At each step $t$, \ourmethod{} predicts the clean sample $\hat{x}_0$, and diffuses it back to $x_{t-1}$. 
}
\label{fig:overview}
\end{figure}

An overview of our method is described in Figure~\ref{fig:overview}. %
Our goal is to synthesize a human motion $x^{1:N}$ of length $N$ given an arbitrary condition $c$. This condition can be any real-world signal that will dictate the synthesis, such as audio~\citep{li2021dance,aristidou2022rhythm}, natural language (text-to-motion)~\citep{tevet2022motionclip,guo2022generating} or a discrete class (action-to-motion)~\citep{guo2020action2motion,petrovich21actor}. 
In addition, unconditioned motion generation is also possible, which we denote as the null condition $c=\emptyset$.
The generated motion $x^{1:N}=\{x^i\}_{i=1}^{N}$ is a sequences of human poses represented by either joint rotations or positions $x^i\in\R^{J\times D}$, where $J$ is the number of joints and $D$ is the dimension of the joint representation. 
\ourmethod{} can accept motion represented by either locations, rotations, or both (see Section~\ref{sec:experiments}).

\textbf{Framework.} Diffusion is modeled as a Markov noising process, $\{x^{1:N}_{t}\}_{t=0}^T$,  where $x^{1:N}_{0}$ is drawn from the data distribution and
\begin{equation}
{
q(x^{1:N}_{t} | x^{1:N}_{t-1}) = \mathcal{N}(\sqrt{\alpha_{t}}x^{1:N}_{t-1},(1-\alpha_{t})I),
}
\end{equation}
where $\alpha_{t} \in (0,1)$ are constant hyper-parameters. When  $\alpha_{t}$ is small enough, we can approximate $x^{1:N}_{T} \sim \mathcal{N}(0,I)$.
From here on we use $x_{t}$ to denote the full sequence at noising step $t$.

In our context, conditioned motion synthesis models the distribution $p(x_{0}|c)$  as the reversed diffusion process of gradually cleaning $x_{T}$. Instead of predicting $\epsilon_t$ as formulated by \citet{ho2020denoising}, we follow \citet{ramesh2022hierarchical} and predict the signal itself, i.e.,
$\hat{x}_{0} = G(x_t, t, c)$
with the \textit{simple} objective~\citep{ho2020denoising},
\begin{equation}
{
\mathcal{L}_\text{simple} = E_{x_0 \sim q(x_0|c), t \sim [1,T]}[\| x_0 - G(x_t, t, c)\|_2^2]
}
\end{equation}
\vspace{-15pt}

\textbf{Geometric losses.} 
In the motion domain, generative networks are standardly regularized
using geometric losses~\cite{petrovich21actor,shi2020motionet}. These losses enforce physical properties and prevent artifacts, encouraging natural and coherent motion.
In this work we experiment with three common geometric losses that regulate (1) positions (in case we predict rotations),  (2) foot contact, and (3) velocities.

\vspace{-20pt}
\begin{equation}
{
\mathcal{L}_\text{pos} = \frac{1}{N} \sum_{i =1}^{N} \| FK(x^i_0) - FK(\hat{x}^i_0)\|_{2}^{2}, %
}
\end{equation}

\begin{equation}
{
\mathcal{L}_\text{foot} = \frac{1}{N-1} \sum_{i =1}^{N-1} \| (FK(\hat{x}^{i+1}_0) - FK(\hat{x}^i_0)) \cdot f_i\|_{2}^{2},
}
\end{equation}
\vspace{-10pt}

\begin{equation}
{
\mathcal{L}_\text{vel} = \frac{1}{N-1} \sum_{i =1}^{N-1} \| (x^{i+1}_0 - x^i_0) - (\hat{x}^{i+1}_0 - \hat{x}^i_0)\|_{2}^{2}
}
\end{equation}

In case we predict joint rotations, $FK(\cdot)$ denotes the forward kinematic function converting joint rotations into joint positions (otherwise, it denotes the identity function). $f_i \in \{0,1\}^J$ is the binary foot contact mask for each frame $i$. Relevant only to feet, it indicates whether they touch the ground, and are set according to binary ground truth data \citep{shi2020motionet}. In essence, it mitigates the foot-sliding effect by nullifying velocities when touching the ground.

Overall, our training loss is
\begin{equation}
{
\mathcal{L} = \mathcal{L}_\text{simple} + \lambda_\text{pos}\mathcal{L}_\text{pos} + \lambda_\text{vel}\mathcal{L}_\text{vel} + \lambda_\text{foot}\mathcal{L}_\text{foot}.
}
\end{equation}

\textbf{Model.} Our model is illustrated in Figure~\ref{fig:overview}. We implement $G$ with a straightforward transformer~\citep{vaswani2017attention} encoder-only architecture.
The transformer architecture is temporally aware, enabling learning arbitrary length motions, and is well-proven for the motion domain~\citep{petrovich21actor,duan2021single,aksan2021spatio}.
The noise time-step $t$ and the condition code $c$ are each projected to the transformer dimension by separate feed-forward networks, then summed to yield the token $z_{tk}$. Each frame of the noised input $x_t$ is linearly projected into the transformer dimension and summed with a standard positional embedding. $z_{tk}$ and the projected frames are then fed to the encoder.
Excluding the first output token (corresponding to $z_{tk}$), the encoder result is projected back to the original motion dimensions, and serves as the prediction $\hat{x}_{0}$. 
We implement text-to-motion by encoding the text prompt to $c$ with CLIP~\citep{radford2021learning} text encoder, and action-to-motion with learned embeddings per class.

\textbf{Sampling} from $p(x_{0}|c)$ is done in an iterative manner, according to \citet{ho2020denoising}. In every time step $t$ we predict the clean sample $\hat{x}_{0} = G(x_t, t, c)$ and noise it back to $x_{t-1}$. This is repeated from $t=T$ until $x_0$ is achieved (Figure~\ref{fig:overview} right).
We train our model $G$ using classifier-free guidance~\citep{ho2022classifier}. In practice, $G$ learns both the conditioned and the unconditioned distributions by randomly setting $c=\emptyset$ for $10\%$ of the samples, such that $G(x_t, t, \emptyset)$ approximates $p(x_0)$. Then, when sampling $G$ we can trade-off diversity and fidelity by interpolating or even extrapolating the two variants using $s$:
\begin{equation}
{
G_s(x_t, t, c) = G(x_t, t, \emptyset) + s \cdot (G(x_t, t, c) - G(x_t, t, \emptyset))
}
\end{equation}

\textbf{Editing.} We enable motion in-betweening in the temporal domain, and body part editing in the spatial domain, by adapting diffusion inpainting to motion data. Editing is done only during sampling, without any training involved. Given a subset of the motion sequence inputs, when sampling the model (Figure~\ref{fig:overview} right), at each iteration we overwrite $\hat{x}_{0}$ with the input part of the motion. 
This encourages the generation to remain coherent to original input, while completing the missing parts. 
In the temporal setting, the prefix and suffix frames of the motion sequence are the input, and we solve a motion in-betweening problem~\citep{harvey2020robust}. Editing can be done either conditionally or unconditionally (by setting $c=\emptyset$). In the spatial setting, we show that body parts can be re-synthesized according to a condition $c$ while keeping the rest intact, through the use of the same completion technique.

\vspace{-5pt}
\section{Experiments}
\label{sec:experiments}
\vspace{-5pt}

\begin{figure}[t!]
\centering
\begin{overpic}[width=\textwidth]{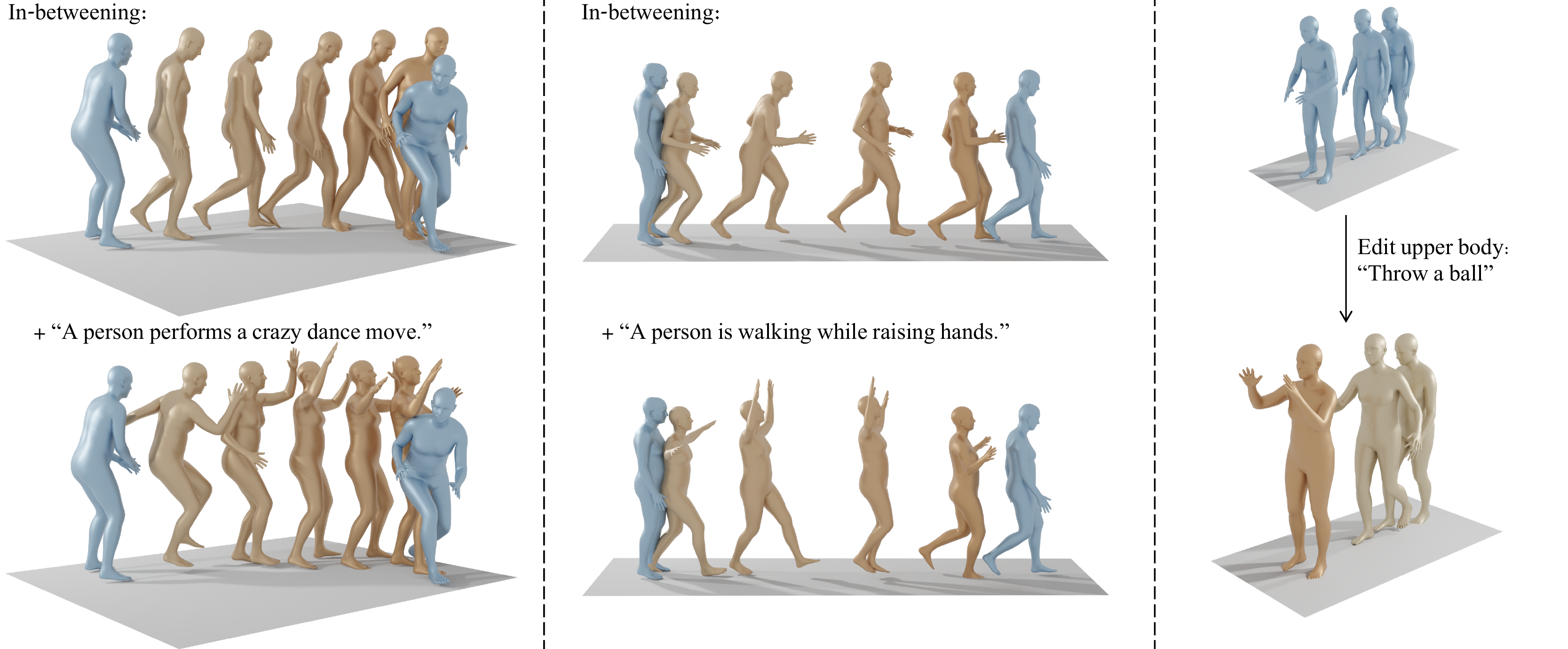}
\end{overpic}
\vspace{-5pt}
\caption{
\textbf{Editing applications.} Light blue frames represent motion input and bronze frames are the generated motion. Motion in-betweening (left+center) can be performed conditioned on text or without condition by the same model. Specific body part editing using text is demonstrated on the right: the lower body joints are fixed to the input motion while the upper body is altered to fit the input text prompt. 
}
\vspace{-10pt}
\label{fig:edit}
\end{figure}

We implement \ourmethod{} for three motion generation tasks: Text-to-Motion(\ref{sec:t2m}), Action-to-Motion(\ref{sec:a2m}) and unconditioned generation(\ref{sec:uncond}. Each sub-section reviews the data and metrics of the used benchmarks, provides implementation details, and presents qualitative and quantitative results. Then, we show implementations of motion in-betweening (both conditioned and unconditioned) and body-part editing by adapting diffusion inpainting to motion (\ref{sec:edit}).
Our models have been trained with $T=1000$ noising steps and a cosine noise schedule.
All of them have been trained on a single \textit{NVIDIA GeForce RTX 2080 Ti} GPU for a period of about $3$ days.

\vspace{-5pt}
\subsection{Text-to-Motion}
\label{sec:t2m}
\vspace{-5pt}

Text-to-motion is the task of generating motion given an input text prompt. The output motion is expected to be both implementing the textual description, and a valid sample from the data distribution (i.e. adhering to general human abilities and the rules of physics). In addition, for each text prompt, we also expect a distribution of motions matching it, rather than just a single result. We evaluate our model using two leading benchmarks - KIT~\citep{plappert2016kit} and HumanML3D~\citep{guo2022generating}, over the set of metrics suggested by \citet{guo2022generating}: 
 \textit{R-precision} and \textit{Multimodal-Dist} measure the relevancy of the generated motions to the input prompts, \textit{FID} measures the dissimilarity between the generated and ground truth distributions (in latent space), \textit{Diversity} measures the variability in the resulting motion distribution, and \textit{MultiModality} is the average variance given a single text prompt. For the full implementation of the metrics, please refer to \citet{guo2022generating}.
We use HumanML3D as a platform to compare different backbones of our model, discovering that the diffusion framework is relatively agnostic to this attribute.
In addition, we conduct a user study comparing our model to current art and ground truth motions. 

\textbf{Data.}
HumanML3D is a recent dataset, textually re-annotating motion capture from the AMASS~\citep{AMASS:ICCV:2019} and HumanAct12~\citep{guo2020action2motion} collections. 
It contains $14,616$ motions annotated by $44,970$ textual descriptions.
In addition, it suggests a redundant data representation including a concatenation of root velocity, joint positions, joint velocities, joint rotations and the foot contact binary labels. We also use in this section the same representation for the KIT dataset, brought by the same publishers. 
Although limited in the number ($3,911$) and the diversity of samples, most of the text-to-motion research is based on KIT, hence we view it as important to evaluate using it as well. 

\begin{figure}%
    \centering
    \subfloat[\centering KIT User Study]{{\includegraphics[width=5cm]{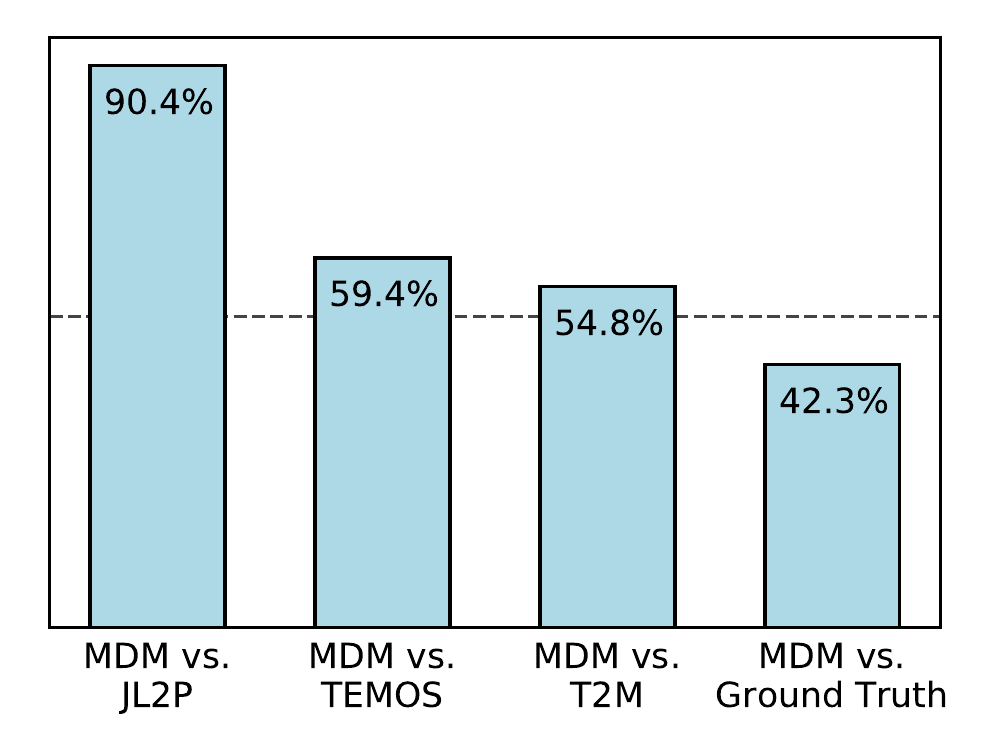} }}%
    \qquad
    \subfloat[\centering Classifier-free scale sweep]{{\includegraphics[width=5cm]{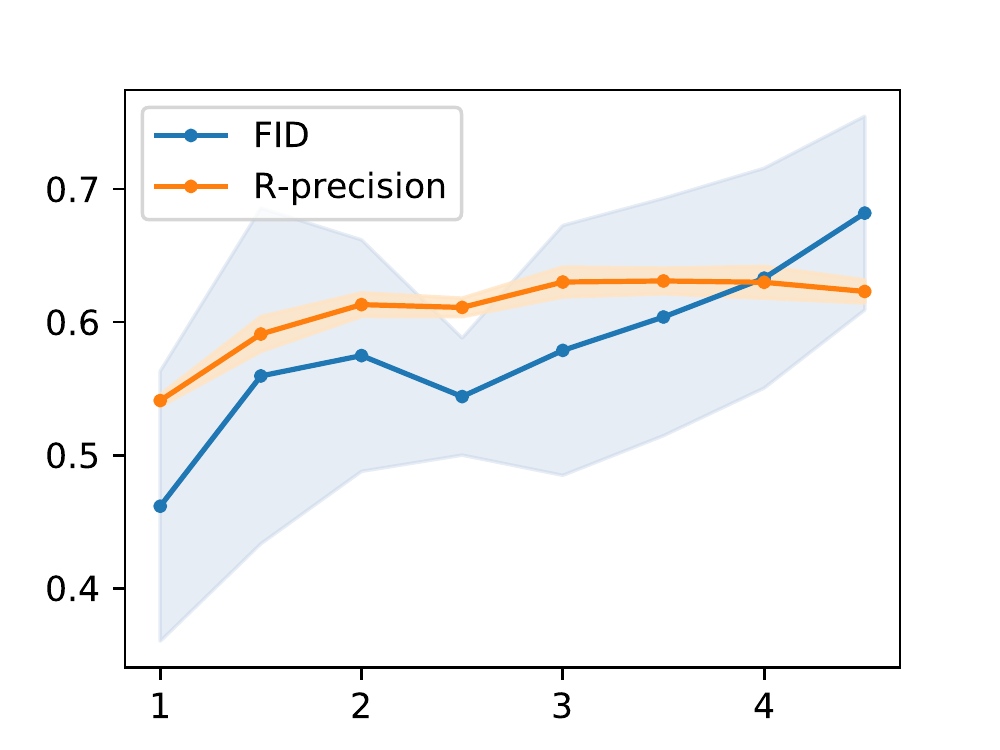} }}%
    \caption{\textbf{(a) Text-to-motion user study for the KIT dataset}. Each bar represents the preference rate of \ourmethod{} over the compared model. \ourmethod{} was preferred over the other models in most of the time, and $42.3\%$ of the cases even over ground truth samples. The dashed line marks $50\%$. \textbf{(b) Guidance-scale sweep for HumanML3D dataset.} \textit{FID} (lower is better) and \textit{R-precision} (higher is better) metrics as a function of the scale $s$, draws an accuracy-fidelity sweet spot around $s=2.5$.}
\vspace{-10pt}
    \label{fig:user_study_and_sweep}
\end{figure}

\textbf{Implementation.}
In addition to our Transformer encoder-only backbone (Section~\ref{sec:method}), we experiment \ourmethod{} with three more backbones: (1) \textit{Transformer decoder} injects $z_{tk}$ through the cross-attention layer, instead of as an input token. (2) \textit{Transformer decoder + input token}, where $z_{tk}$ is injected both ways, and (3) \textit{GRU}~\citep{cho2014learning} concatenate $z_{tk}$ to each input frame (Table~\ref{tab:humanml3d}). 
Our models were trained with batch size $64$, $8$ layers (except GRU that was optimal at $2$), and latent dimension $512$. To encode the text we use a frozen \textit{CLIP-ViT-B/32} model.
Each model was trained for $500K$ steps, afterwhich a checkpoint was chosen that minimizes the FID metric to be reported. Since foot contact and joint locations are explicitly represented in HumanML3D, we don't apply geometric losses in this section. We evaluate our models with guidance-scale $s=2.5$ which provides a diversity-fidelity sweet spot (Figure~\ref{fig:user_study_and_sweep}).

\textbf{Quantitative evaluation.} We evaluate and compare our models to current art (JL2P~\citet{ahuja2019language2pose}, Text2Gesture~\citep{bhattacharya2021text2gestures}, and T2M~\citep{guo2022generating}) with the metrics suggested by \citet{guo2022generating}. 
As can be seen, \ourmethod{} achieves state-of-the-art results in \textit{FID}, \textit{Diversity}, and \textit{MultiModality}, indicating high diversity per input text prompt, and high-quality samples, as can also be seen qualitatively in Figure~\ref{fig:teaser}.

\textbf{User study.} We asked 31 users to choose between \ourmethod{} and 
state-of-the-art works
in a side-by-side view, with both samples generated from the same text prompt randomly sampled from the KIT test set. We repeated this process with 10 samples per model and 10 repetitions per sample. 
This user study enabled a comparison with the recent TEMOS model~\citep{petrovich22temos}, which was not included in the HumanML3D benchmark.
Fig.~\ref{fig:user_study_and_sweep} shows that most of the time, \ourmethod{} was preferred over the compared models, and even preferred over ground truth samples in $42.3\%$ of the cases.

\begin{table}[ht]
\centering
\resizebox{\textwidth}{!}{

\begin{tabular}{lccccc}
\toprule

\multirow{2}{2cm}{\centering Method} & 
\multirow{2}{2.cm}{\centering R Precision (top 3)$\uparrow$} & 
\multirow{2}{1.5cm}{\centering FID$\downarrow$} & \multirow{2}{2.5cm}{\centering Multimodal Dist$\downarrow$} & \multirow{2}{2cm}{\centering Diversity$\rightarrow$} & \multirow{2}{2cm}{\centering Multimodality$\uparrow$} \\
\\

\midrule
Real & $0.797^{\pm.002}$ & $0.002^{\pm.000}$ & $2.974^{\pm.008}$ & $9.503^{\pm.065}$ & -\\ 
\midrule

JL2P &  $0.486^{\pm.002}$ & $11.02^{\pm.046}$ & $5.296^{\pm.008}$ & $7.676^{\pm.058}$ & - \\

Text2Gesture &  $0.345^{\pm.002}$ & $7.664^{\pm.030}$ & $6.030^{\pm.008}$ & $6.409^{\pm.071}$ & - \\

T2M & $\mathbf{0.740^{\pm.003}}$ & $1.067^{\pm.002}$ & $\mathbf{3.340^{\pm.008}}$ & $9.188^{\pm.002}$ & $2.090^{\pm.083}$ \\

\midrule

\ourmethod{} (ours) &  ${0.611^{\pm.007}}$ & $\mathbf{0.544^{\pm.044}}$ & ${5.566^{\pm.027}}$ & $\mathbf{9.559^{\pm.086}}$ & ${2.799^{\pm.072}}$ \\

\ourmethod{} (decoder) & ${0.608^{\pm.005}}$ & ${0.767^{\pm.085}}$ & ${5.507^{\pm.020}}$ & ${9.176^{\pm.070}}$ & $\mathbf{2.927^{\pm.125}}$ \\

 \,\,\,\,\, + input token & ${0.621^{\pm.005}}$ & ${0.567^{\pm.051}}$ & ${5.424^{\pm.022}}$ & ${9.425^{\pm.060}}$ & $2.834^{\pm.095}$ \\

\ourmethod{} (GRU) & ${0.645^{\pm.005}}$ & ${4.569^{\pm.150}}$ & ${5.325^{\pm.026}}$ & ${7.688^{\pm.082}}$ & $1.2646^{\pm.024}$ \\

\bottomrule
\end{tabular}
}
\caption{\textbf{Quantitative results on the HumanML3D test set.} All methods use the real motion length from the ground truth. `$\rightarrow$' means results are better if the metric is closer to the real distribution. We run all the evaluation 20 times (except \textit{MultiModality} runs 5 times) and $\pm$ indicates the 95\% confidence interval. \textbf{Bold} indicates best result.}
\label{tab:humanml3d}
\end{table}

\begin{table*}[ht]
\centering

\resizebox{\textwidth}{!}{

\begin{tabular}{lccccc}
\toprule

\multirow{2}{2cm}{\centering Method} & 
\multirow{2}{2.cm}{\centering R Precision (top 3)$\uparrow$} & 
\multirow{2}{1.5cm}{\centering FID$\downarrow$} & \multirow{2}{2.5cm}{\centering Multimodal Dist$\downarrow$} & \multirow{2}{2cm}{\centering Diversity$\rightarrow$} & \multirow{2}{2cm}{\centering Multimodality$\uparrow$} \\
\\

\midrule
Real & $0.779^{\pm.006}$ & $0.031^{\pm.004}$ & $2.788^{\pm.012}$ & $11.08^{\pm.097}$ & -\\ 
\midrule

JL2P & $0.483^{\pm.005}$ & $6.545^{\pm.072}$ & $5.147^{\pm.030}$ & $9.073^{\pm.100}$ & - \\

Text2Gesture & $0.338^{\pm.005}$ & $12.12^{\pm.183}$ & $6.964^{\pm.029}$ & $9.334^{\pm.079}$ & - \\

T2M & $\mathbf{0.693^{\pm.007}}$ & ${2.770^{\pm.109}}$ & $\mathbf{3.401^{\pm.008}}$ & $\mathbf{10.91^{\pm.119}}$ & $1.482^{\pm.065}$ \\
\midrule

\ourmethod{} (ours) & ${0.396^{\pm.004}}$ & $\mathbf{0.497^{\pm.021}}$ & ${9.191^{\pm.022}}$ & ${10.847^{\pm.109}}$ & $\mathbf{1.907^{\pm.214}}$\\

\bottomrule
\end{tabular}
}
\caption{\textbf{Quantitative results on the KIT test set.} %
}
\label{tab:kit}
\end{table*}

\begin{table}[tbh]
\centering

\resizebox{.95\textwidth}{!}{

    \begin{tabular}{lccccc}
    \toprule
    
    \multirow{2}{2cm}{\centering Method} & 
    \multirow{2}{1.5cm}{\centering FID$\downarrow$} &
    \multirow{2}{1.5cm}{\centering Accuracy$\uparrow$} &
    \multirow{2}{2cm}{\centering Diversity$\rightarrow$} &
    \multirow{2}{2cm}{\centering Multimodality$\rightarrow$} \\
    \\
    
    \midrule
    Real (INR) & $0.020^{\pm.010}$ & $0.997^{\pm.001}$ & $6.850^{\pm.050}$ & $2.450^{\pm.040}$\\ 

    Real (ours) & $0.050^{\pm.000}$ & $0.990^{\pm.000}$ & $6.880^{\pm.020}$ & $2.590^{\pm.010}$\\   %

    \midrule
    
    Action2Motion~\citeyearpar{guo2020action2motion} & $0.338^{\pm.015}$ & $0.917^{\pm.003}$ & $6.879^{\pm.066}$ & $\underline{2.511^{\pm.023}}$\\ 
    
    ACTOR~\citeyearpar{petrovich21actor} & $0.120^{\pm.000}$ &$0.955^{\pm.008}$ &$\mathbf{6.840^{\pm.030}}$ &$2.530^{\pm.020}$ \\ 

    INR~\citeyearpar{cervantes2022implicit} &  $\underline{0.088^{\pm.004}}$ & $\underline{0.973^{\pm.001}}$ & $6.881^{\pm.048}$ & $2.569^{\pm.040}$ \\ 
    \midrule

    \ourmethod{} (ours) &
        $0.100^{\pm.000}$ & $\mathbf{0.990^{\pm.000}}$ & $\underline{6.860^{\pm.050}}$ & $2.520^{\pm.010}$\\

    \quad\quad w/o foot contact  & 
        $\mathbf{0.080^{\pm.000}}$ & $\mathbf{0.990^{\pm.000}}$ & $6.810^{\pm.010}$ & $\mathbf{2.580^{\pm.010}}$ \\  %

    \bottomrule
    \end{tabular}
}
\caption{\textbf{Evaluation of action-to-motion on the HumanAct12 dataset.} Our model leads the board in three out of four metrics. 
Ground-truth evaluation results are slightly different for each of the works, due to implementation differences, such as python package versions. 
It is important to assess the diversity and multimodality of each model using its own ground-truth results, as they are measured by their distance from GT. We show the GT metrics measured by our model and by the leading compared work, INR~\citep{cervantes2022implicit}.
\textbf{Bold} indicates best result, $\underline{\text{underline}}$ indicates second best, $\pm$ indicates 95\% confidence interval, $\rightarrow$ indicates that closer to real is better.}
\label{tab:humanact12}
\end{table}

\begin{table}[tbh]
\centering

\resizebox{\textwidth}{!}{

    \begin{tabular}{lccccc}
    \toprule
    
    \multirow{2}{2cm}{\centering Method} & 
    \multirow{2}{1.5cm}{\centering $\text{FID}_{\text{train}}\downarrow$} &
    \multirow{2}{1.5cm}{\centering $\text{FID}_{\text{test}}\downarrow$} &
    \multirow{2}{1.5cm}{\centering Accuracy$\uparrow$} &
    \multirow{2}{2cm}{\centering Diversity$\rightarrow$} &
    \multirow{2}{2cm}{\centering Multimodality$\rightarrow$} \\
    \\
    
    \midrule
    Real & $2.92^{\pm.26}$ & $2.79^{\pm.29}$ & $0.988^{\pm.001}$ & $33.34^{\pm.320}$ & $14.16^{\pm.06}$\\ 
    \midrule
    
    ACTOR~\citeyearpar{petrovich21actor} & $20.49^{\pm2.31}$ & $23.43^{\pm2.20} $ &$0.911^{\pm.003}$ & $31.96^{\pm.33}$ & $14.52^{\pm.09}$ \\ 

    INR~\citeyearpar{cervantes2022implicit} (best variation) &  $\mathbf{9.55^{\pm.06}}$ & $15.00^{\pm.09}$ &$0.941^{\pm.001}$ & $31.59^{\pm.19}$ & $14.68^{\pm.07}$ \\ 
    \midrule

    \ourmethod{} (ours)  & 
        $9.98^{\pm1.33}$ & $\mathbf{12.81^{\pm1.46}}$ & $\underline{0.950^{\pm.000}}$ & $\underline{33.02^{\pm.28}}$ & $\mathbf{14.26^{\pm.12}}$ \\

    \quad\quad w/o foot contact & 
        $\underline{9.69^{\pm.81}}$ & $\underline{13.08^{\pm2.32}}$ & $\mathbf{0.960^{\pm.000}}$ & $\mathbf{33.10^{\pm.29}}$ & $\mathbf{14.06^{\pm.05}}$\\

    \bottomrule
    \end{tabular}

} %
\caption{\textbf{Evaluation of action-to-motion on the UESTC dataset.} The performance improvement with our 
model shows a clear gap from state-of-the-art. 
\textbf{Bold} indicates best result, $\underline{\text{underline}}$ indicates second best, $\pm$ indicates 95\% confidence interval, $\rightarrow$ indicates that closer to real is better.}
\label{tab:uestc}
\end{table}
\vspace{-10pt}

\subsection{Action-to-Motion}
\vspace{-5pt}

\label{sec:a2m}
Action-to-motion is the task of generating motion given an input action class, represented by a scalar. The output motion should faithfully animate the input action, and at the same time be natural and reflect the distribution of the dataset on which the model is trained. Two dataset are commonly used to evaluate action-to-motion models: HumanAct12~\citep{guo2020action2motion} and UESTC~\citep{ji2018large}. We evaluate our model using the set of metrics suggested by \citet{guo2020action2motion}, namely Fr\'{e}chet Inception Distance (FID), action recognition accuracy, diversity and multimodality. The combination of these metrics makes a good measure of the realism and diversity of generated motions.

\textbf{Data.}
HumanAct12~\citep{guo2020action2motion} offers approximately 1200 motion clips, organized into 12 action categories, with 47 to 218 samples per label. 
UESTC~\citep{ji2018large} consists of 40 action classes, 40 subjects and 25K samples, and is split to train and test. 
We adhere to the cross-subject testing protocol used by current works, with 225-345 samples per action class.
For both datasets we use the sequences provided by \citet{petrovich21actor}.

\textbf{Implementation.}
The implementation presented in Figure~\ref{fig:overview} holds for all the variations of our work. In the case of action-to-motion, the only change would be the substitution of the text embedding by an action embedding. Since action is represented by a scalar, its embedding is fairly simple; each input action class scalar is converted into a learned embedding of the transformer dimension.

The experiments have been run with batch size $64$, a latent dimension of $512$, and an encoder-transformer architecture. Training on HumanAct12 and UESTC has been carried out for $750K$ and $2M$ steps respectively. In our tables we display the evaluation of the checkpoint that minimizes the FID metric. 

\textbf{Quantitative evaluation.}
Tables \ref{tab:humanact12} and \ref{tab:uestc} reflect \ourmethod{}'s performance on the HumanAct12 and UESTC datasets respectively. We conduct 20 evaluations, with 1000 samples in each, and report their average and a 95\% confidence interval.
We test two variations, with and without foot contact loss.
Our model leads the board for both datasets. 
The variation with no foot contact loss attains slightly better results; nevertheless, as shown in our supplementary video, the contribution of foot contact loss to the quality of results is important, and without it we witness artifacts such as shakiness and unnatural gestures.

\vspace{-5pt}
\section{Additional Applications}
\vspace{-5pt}
\subsection{Motion Editing} \label{sec:edit}
\vspace{-5pt}

In this section we implement two motion editing applications - \textbf{in-betweening} and \textbf{body part editing}, both using the same approach in the temporal and spatial domains correspondingly. 
For \textbf{in-betweening}, we fix the first and last $25\%$ of the motion, leaving the model to generate the remaining $50\%$ in the middle. For \textbf{body part editing}, we fix the joints we don't want to edit and leave the model to generate the rest. In particular, we experiment with editing the upper body joints only.
In figure~\ref{fig:edit} we show that in both cases, using the method described in Section~\ref{sec:method} generates smooth motions that adhere both to the fixed part of the motion and the condition (if one was given).

\begin{table}[tbh]
\centering

\small

    \begin{tabular}{lcccccc}
    \toprule
    
    \multirow{2}{2cm}{\centering Method} & 
    \multirow{2}{1.5cm}{\centering FID$\downarrow$} &
    \multirow{2}{2cm}{\centering KID$\downarrow$} &
    \multirow{2}{2cm}{\centering Precision$\uparrow$ Recall$\uparrow$} &
    \multirow{2}{2cm}{\centering Multimodality$\uparrow$} \\
    \\

    \midrule

    ACTOR~\citeyearpar{petrovich21actor} & 48.80 & 0.53 & 0.72,   0.74 & 14.10 \\ 

    MoDi~\citeyearpar{raab2022modi} & \textbf{13.03} & \textbf{0.12} & \textbf{0.71},   \textbf{0.81} & \textbf{17.57}\\ 
    
    \midrule

    \ourmethod{} (ours)  & 31.92 & 0.36 & 0.66, 0.62 & 17.00\\  

    \bottomrule
    \end{tabular}
\caption{\textbf{Evaluation of unconstrained synthesis on the HumanAct12 dataset.} We test \ourmethod{} in the challenging unconstrained setting, and compare with MoDi~\citep{raab2022modi}, a work that was specially designed for such setting. We demonstrate that in addition to being able to support any condition, we can achieve plausible results in the unconstrained setting.
\textbf{Bold} indicates best result.}
\label{tab:unconstrained}
\end{table}

\vspace{-5pt}
\subsection{Unconstrained Synthesis} \label{sec:uncond}
\vspace{-5pt}
The challenging task of unconstrained synthesis has been studied by only a few~\citep{holden2016deep, raab2022modi}. 
In the presence of data labeling, e.g., action classes or text description, the labels work as a supervising factor, and facilitate a structured latent space for the training network. 
The lack of labeling make training more difficult.
The human motion field possesses rich unlabeled datasets~\citep{mixamo}, and the ability to train on top of them is an advantage.
Daring to test \ourmethod{} in the challenging unconstrained setting, we follow MoDi\citep{raab2022modi} for evaluation. We use the metrics they suggest (FID, KID, precision/recall and multimodality), and run on an unconstrained version of the HumanAct12~\citep{guo2020action2motion} dataset.

\textbf{Data.}
Although annotated, we use HumanAct12 (see Section \ref{sec:a2m}) in an unconstrained fashion, ignoring its labels. The choice of HumanAct12 rather than a dataset with no labels (e.g., Mixamo~\citep{mixamo}), is for compatibility with previous publications. 

\textbf{Implementation.}
Our model uses the same architecture for all forms of conditioning, as well as for the unconstrained setting.
The only change to the structure shown in Figure~\ref{fig:overview}, is the removal of the conditional input, such that $z_{tk}$ is composed of the projection of $t$ only.
To simulate an unconstrained behavior, ACTOR~\citet{petrovich21actor} has been trained by \citep{raab2022modi} with a labeling of one class to all motions.

\textbf{Quantitative evaluation.}
The results of our evaluation are shown in table \ref{tab:unconstrained}.
We demonstrate superiority over works that were not designed for an unconstrained setting, and get closer to MoDi~\citep{raab2022modi}. MoDi is carefully molded for unconstrained settings, while our work can be applied to any (or no) constrain, and also provides editing capabilities.

\vspace{-5pt}
\section{Discussion}
\vspace{-5pt}

We have presented \ourmethod{}, a method that lends itself to various human motion generation tasks. \ourmethod{} is an untypical classifier-free diffusion model, featuring a transformer-encoder backbone, and predicting the signal, rather than the noise.
This yields both a lightweight model, that is unburdening to train, and an accurate one, gaining much from the applicable geometric losses. Our experiments show superiority in conditioned generation, but also that this approach is not very sensitive to the choice of architecture. 

A notable limitation of the diffusion approach is the long inference time, requiring about $1000$ forward passes for a single result. Since our motion model is small  anyway, using dimensions order of magnitude smaller than images, our inference time shifts from less than a second to only about a minute, which is an acceptable compromise.  
As diffusion models continue to evolve, beside better compute, in the future we would be interested in seeing how to incorporate better control into the generation process, and widen the options for applications even further.

\section*{Acknowledgements}
We thank Rinon Gal for his useful suggestions and references.
This research was supported in part by the Israel Science Foundation (grants no. 2492/20 and 3441/21), Len Blavatnik and the Blavatnik family foundation, and The Tel Aviv University Innovation Laboratories (TILabs).

\bibliography{mdm}
\bibliographystyle{iclr2023_conference}

\appendix

\end{document}